\documentclass{article}

\usepackage[english]{babel}
\usepackage{algorithm}
\usepackage{algorithmic}
\usepackage[justification=centering]{caption}
\usepackage[letterpaper,top=2cm,bottom=2cm,left=3cm,right=3cm,marginparwidth=1.75cm]{geometry}

\usepackage{amsmath}
\usepackage{amssymb} 
\usepackage{graphicx}
\usepackage[colorlinks=true, allcolors=blue]{hyperref}

\usepackage{amsmath}
\usepackage{amsfonts}
\usepackage{amssymb}
\usepackage[utf8]{inputenc}
\usepackage{amsmath}
\usepackage{amsfonts}
\usepackage{graphicx,color}
\usepackage{hyperref}
\usepackage{booktabs}
\usepackage[super]{nth}
\hypersetup{
  colorlinks=TRUE,
  linkcolor=blue,
  urlcolor=blue,
  citecolor=blue
}
\usepackage{authblk}
\usepackage{epsfig}
\usepackage{amssymb}
\usepackage{fancyhdr}
\usepackage{amsmath}
\usepackage{multirow}
\usepackage{algorithm}
\usepackage{geometry}

\usepackage{algorithm}
\usepackage{algorithmic}
\usepackage{float}

\title{ On weight and variance uncertainty in neural networks
for regression tasks}

\author[1]{Moein Monemi\thanks{moein.monemi@ut.ac.ir}}
\author[2]{Morteza Amini\thanks{Corresponding author, e-mail: morteza.amini@ut.ac.ir}}
\author[1]{S. Mahmoud Taheri\thanks{sm\_taheri@ut.ac.ir}}
\author[3]{Mohammad Arashi\thanks{arashi@um.ac.ir}}
\affil[1]{School of Engineering Science, College of Engineering, University of Tehran, Tehran, Iran}
\affil[2]{Department of Statistics, School of Mathematics, Statistics, and Computer Science, College of Science, University of Tehran, Tehran, Iran}
\affil[3]{Department of Statistics, Faculty of Mathematical Sciences, Ferdowsi University of Mashhad, P.O. Box 1159, Mashhad 91775, Iran}

\begin{document}
\maketitle

\begin{abstract}
 We investigate the problem of weight uncertainty originally proposed by [Blundell et al. (2015). Weight uncertainty in neural networks. In International conference on machine learning, 1613-1622, PMLR.] in the context of neural networks designed for regression tasks, and we extend their framework by incorporating variance uncertainty into the model. Our analysis demonstrates that explicitly modeling uncertainty in the variance parameter can significantly enhance the predictive performance of Bayesian neural networks. By considering a full posterior distribution over the variance, the model achieves improved generalization compared to approaches that treat variance as fixed or deterministic. We evaluate the generalization capability of our proposed approach through a function approximation example and further validate it on the riboflavin genetic dataset. Our exploration encompasses both fully connected dense networks and dropout neural networks, employing Gaussian and spike-and-slab priors respectively for the network weights, providing a comprehensive assessment of how variance uncertainty affects model performance across different architectural choices.
\end{abstract}


\section{Introduction}
Bayesian Neural Networks (BNNs) have been introduced and comprehensively discussed by many authors (among others see \cite{Neal1992,Bishop1997}). BNNs are suitable for modeling uncertainty by considering values of the parameters that might not be learned by the available data. This is achieved by randomizing unknown parameters, such as weights and biases while incorporating prior knowledge. Such an approach naturally regularizes the model and helps prevent overfitting, a common challenge in neural networks. \textcolor{black}{Recently, other approaches such as heuristic shrinkage penalized methods have also been proposed to address the regression problems in deep neural networks \cite{Behzadi2025}.} The BNNs can also be considered as a suitable and more reliable alternative to ensemble learning methods \cite{Price2023}, including bagging and boosting of the neural networks.

The primary objective in {the BNNs} is to estimate the posterior distribution of the parameters. However, obtaining the exact posterior is often infeasible due to the integrals' intractability. Thus, it is necessary to use approximation techniques for the posterior density. One common approach is the Markov Chain Monte Carlo (MCMC), including the Metropolis-Hastings algorithm \cite{Hastings1970}, which generates samples from the posterior distribution by constructing a Markov chain, typically of first orders. 

Despite its effectiveness, flexibility, and strong theoretical foundations, MCMC has significant limitations in high-dimensional spaces, including slow convergence and high computational costs. Variational Bayes (VB) has emerged as a faster and more scalable approximation technique. First introduced by \cite{Jordan1999} and \cite{Wainwright2008}, VB seeks to approximate the intractable posterior distribution by identifying a simpler, parametric probability distribution that closely resembles the true posterior. The central idea in VB is to recast the problem of posterior inference as an optimization task. Specifically, VB minimizes the Kullback-Leibler (KL) divergence \cite{Kullback1951} between the approximate distribution and the true posterior \cite{Blei2017}. This involves iteratively updating the hyper-parameters of the approximate distribution to improve its fit to the posterior while maintaining computational efficiency \cite{Ahmed2012}. 

The VB method has limitations related to the conjugacy of the full conditionals of parameters. In such cases, models that go beyond the conjugacy of the exponential family have been proposed \cite{Zhang2018}. Black Box Variational Inference (BBVI) \cite{Ranganath2014} eliminates the need for conditional conjugacy by introducing a flexible framework for approximating posterior distributions. It leverages stochastic optimization techniques to estimate gradients and employs variance reduction strategies to enhance computational efficiency and stability. Bayes by Backprop (weight uncertainty), introduced by Blundell et al. \cite{Blundell2015}, is another approach for performing Bayesian inference in {NNs}. This method leverages stochastic optimization and employs the re-parametrization trick to enable efficient sampling and gradient-based optimization of posterior distributions, {for a NN} weights. The model is used for classification and regression tasks, by considering suitable multi-nomial and Gaussian likelihood for the target variable, respectively.

{ Another approach for modeling uncertainty in weights of a NN is considered in \cite{galg16}. They modeled the variational approximation of the weights as the multiplication of a fixed hyperparameter (which might be optimized) by a bernoulli random varaible for modeling the dropout mechanism. There are two main differences between the method considered in \cite{galg16} and that in \cite{Blundell2015}. The first difference is that no statistical distribution is considered for modelling uncertainty in weights around the hyperparameter in the method proposed in \cite{galg16}. The second difference is the approach for modeling the dropout mechanism, since the dropout mechanism was modeled by considering the slab-and-spike prior for the weights of the NN in \cite{Blundell2015}. }

For the regression task, the variance of the Gaussian likelihood is assumed fixed in \cite{Blundell2015} and is determined by cross-validation. In this paper, we consider the problem of variance uncertainty in the model proposed by \cite{Blundell2015} to investigate its effectiveness on the prediction performance of the BNNs for regression tasks. Throughout a simulation study from a nonlinear function approximation problem, it is shown that the Bayes by Backprop with variance uncertainty performs better than the model with fixed variance. We have also considered the riboflavin data set, which is a genetic high dimensional data set to evaluate the performance of the proposed model, compared to the other methods, by applying PCA-BNN and the dropout-BNN methods. The dropout-BNN is applied by considering the spike-and-slab prior for the weights of the BNN. As the results presented in Section 6 shows, our proposed method outperforms other approaches in terms of MSPE and coverage probabilities for the regression curve, as well as MSPE for the riboflavin dataset.

\textcolor{black}{
While weight uncertainty has been extensively studied, treating the likelihood variance as a random variable in BNNs offers distinct advantages that are often overlooked. We distinguish our approach from heteroscedastic regression models (e.g., \cite{Kendall2017}), where the network predicts a data-dependent variance $\sigma(x)$ for each input. In contrast, our work focuses on the \textit{epistemic uncertainty} of the global observation variance. This is particularly important in limited-data problems, where the true variance is unknown and assuming a fixed or point-estimated variance can lead to overconfident predictions.
Furthermore, unlike classical Bayesian models which typically employ an Inverse-Gamma conjugate prior for the variance, such closed-form updates are intractable in deep neural networks. Our proposed method extends the Bayes-by-Backprop framework to the likelihood variance, enabling joint optimization of weights and variance parameters via stochastic gradient descent without requiring conjugacy. Finally, compared to optimizing $\sigma^2$ as a point estimate, marginalizing over the variance posterior introduces heavy-tailed behavior in the predictive distribution. This results in a model that is more robust to outliers and provides better prediction intervals, as demonstrated by our coverage probability results in Section 6.
}

The remainder of the paper is organized as follows. Section 2 {reviews} the foundations of the BNNs. The VB method for posterior approximation is introduced in Section 3. Section 4 describes the Bayes by {Backprop method} proposed by \cite{Blundell2015}, for the regression task with fixed variance. In Section 5, we propose the variance uncertainty for the BNNs. The experimental results and the evaluation of the proposed model is considered in Section 6. Some concluding remarks are given in Section 7. The codes for this study is available online on github (soon).

\section{Bayesian Neural Networks}

{In this section, our goal is to describe BNNs and their challenges in practical applications}. Before {examining} BNNs, we briefly describe {NNs} from a statistical perspective. {A NN} aims to approximate the unknown function $\mathbf{y} = {\phi(\mathbf{x})}$ where ${\phi(\mathbf{x})}$ and $\mathbf{y}$ are the input and the output vectors, respectively. { From a statistical point of view, {given the sample points $\mathbf{x}, \mathbf{y}$}, the objective function of NN is the log-likelihood function of weights and biases, $\mathbf{W}$, and extra parameters $\rho$, of the model, $\log L(\mathbf{W},\rho|\mathbf{x},\mathbf{y})$ which is aimed to be maximized to learn the parameters of the network
\[
\widehat{(\mathbf{W},\rho)}_{\text{\rm MLE}} = \arg\max_{(\mathbf{W},\rho)} \log L(\mathbf{W},\rho \mid \mathbf{x},\mathbf{y}).
\]

The likelihood function $L(\mathbf{W},\rho \mid \mathbf{x},\mathbf{y})$ depends on the specific problem being addressed:
\begin{itemize}
    \item In regression tasks, the likelihood function is typically modeled by a Gaussian density of $\mathbf{y}$ with a mean provided by the network given the input {data set} $\mathbf{x}$ and an unknown variance $\sigma^2$, and the negative log-likelihood corresponds to the sum of squared errors (SSE).

    \item In classification problems, the likelihood function is often modeled as a categorical distribution, and the negative log-likelihood corresponds to the cross-entropy loss.
\end{itemize}
In a BNN, a prior distribution $p(\mathbf{W},\rho)$ is considered for the parameters of the model \cite{Neal1992}, and our goal is to find a posterior distribution of all parameters $p(\mathbf{W},\rho \mid \mathbf{x},\mathbf{y})$.
Using the Bayes' formula, the posterior density is obtained as 
\[
p(\mathbf{W},\rho \mid \mathbf{x},\mathbf{y}) = \frac{L(\mathbf{W},\rho \mid \mathbf{x},\mathbf{y}) \, p(\mathbf{W},\rho)}{p(\mathbf{x},\mathbf{y})},
\]
where $p(\mathbf{x},\mathbf{y})$ is the marginal distribution of the data, given by 
$$p(x,y) = \int L(\mathbf{W},\rho \mid \mathbf{x},\mathbf{y}) \, p(\mathbf{W},\rho) \, d(\mathbf{W},\rho).$$
The point estimation of the model parameters might be computed by the maximum a posteriori (MAP) estimator as follows
\[
\widehat{(\mathbf{W},\rho)}_{\text{\rm MAP}} = \arg\max_{(\mathbf{W},\rho)} \log L(\mathbf{W},\rho \mid \mathbf{x},\mathbf{y}) + \log p(\mathbf{W},\rho).
\]
The posterior predictive distribution function, used to make predictions and to determine prediction intervals for new data, is defined as follows ({for more details, see e.g., \cite{Neal1992,Bishop1997}})
\[
p(\mathbf{y^{*}} \mid \mathbf{x},\mathbf{y},\mathbf{x^*}) = \int_{\tilde{\boldsymbol{\theta}}} L(\mathbf{W},\rho \mid \mathbf{x^*},\mathbf{y^*}) \, p(\mathbf{W},\rho \mid \mathbf{x},\mathbf{y}) \, d(\mathbf{W},\rho).
\]
In practice, we face intractable integrals to solve \( p(\mathbf{x},\mathbf{y}) \); therefore, we resort to approximation methods. In the following, we introduce VB as a well-known approximation technique.
}
\section{Variational Bayes}
The VB is a method for finding an approximate distribution $q(\boldsymbol{\theta})$ of the posterior distribution $p(\boldsymbol{\theta} |\mathbf{x})$, by minimizing the Kullback-Leibler divergence ${\rm KL} \left[ q(\boldsymbol{\theta}) || p(\boldsymbol{\theta}|\mathbf{x}) \right]$ as a measure of closeness \cite{tr23}. In this section, we briefly {review the basic elements of} the VB method.

Let $\mathbf{x}$ be a vector of observed data, and $\boldsymbol{\theta}$ be a parameter vector with joint distribution $p(\mathbf{x} , \boldsymbol{\theta})$.
In the Bayesian inference framework, the inference about $\boldsymbol{\theta}$ is done based on the posterior distribution
$p(\boldsymbol{\theta} |\mathbf{x}) = p(\mathbf{x} , \boldsymbol{\theta})/p(\mathbf{x})$
where 
$p(\mathbf{x}) = \int  p(\mathbf{x} , \boldsymbol{\theta}) d\boldsymbol{\theta}$.

In the VB method, we assume the parameter vector $\boldsymbol{\theta}$ is divided into $M$ partitions  
$ \{ \boldsymbol{\theta}_1, \ldots , \boldsymbol{\theta}_M \} $ and we want to approximate
$p(\boldsymbol{\theta} |\mathbf{x}) $
by 
$$q(\boldsymbol{\theta}) = \prod_{j=1}^{M} q_j(\boldsymbol{\theta}_j),$$
where $q_j(.)$ is the variational posterior over the vector of posterior parameters $\boldsymbol{\theta}_j$. Hence, each $\boldsymbol{\theta}_j$ can be a vector of parameters.
The best VB approximation 
$q^{\ast}(\cdot)$
is then obtained as 
$$q^{\ast} = \arg \min_q   {\rm KL} \left[ q(\boldsymbol{\theta}) || p(\boldsymbol{\theta}|\mathbf{x}) \right],$$
where
\begin{equation}\label{kl}
{\rm KL} \left[ q(\boldsymbol{\theta}) || p(\boldsymbol{\theta}|\mathbf{x}) \right] = \int q(\boldsymbol{\theta}) \log \frac{q(\boldsymbol{\theta})}{p(\boldsymbol{\theta}| \mathbf{x})} d\boldsymbol{\theta} =   \log p(\mathbf{x}) - \int q(\boldsymbol{\theta}) \log \frac{p(\mathbf{x},\boldsymbol{\theta})}{q(\boldsymbol{\theta})} d\boldsymbol{\theta}. 
\end{equation}
Let $ \boldsymbol{\eta} = \{ \boldsymbol{\eta}_1, \ldots , \boldsymbol{\eta}_M \} $ be the partitions of hyper-parameter such that each $\boldsymbol{\eta}_j$ be the variational parameter of $\boldsymbol{\theta}_j$. In the fixed-form-VB (FFVB), we consider an assumed density function for each $q(\boldsymbol{\theta}_j|\boldsymbol{\eta}_j)$, with unknown $\boldsymbol{\eta}_j$, and we aim to determine the hyper-parameters $\boldsymbol{\eta}$ such that ${\rm KL} \left[ q(\boldsymbol{\theta}|\boldsymbol{\eta}) || p(\boldsymbol{\theta}|x) \right]$ is minimized.

It is clear from \eqref{kl} that minimizing KL is equivalent to maximizing the evidence lower bound (ELBO) with respect to $\boldsymbol{\eta}$, defined as 
$$ {\rm ELBO} =  \int q(\boldsymbol{\theta}|\boldsymbol{\eta}) \log \frac{p(\mathbf{x},\boldsymbol{\theta})}{q(\boldsymbol{\theta}|\boldsymbol{\eta})} d\boldsymbol{\theta}.$$
The ELBO is equal to $\log p(x)$ when the KL divergence is zero, meaning a perfect fit.
The closer the value of ${\rm ELBO}[q(\boldsymbol{\theta})] $ to $\log p(\mathbf{x})$ (greater values), the fit is better. Maximizing the ELBO is equivalent to minimizing its negative. It follows that
\begin{equation*}
\boldsymbol{\eta}^* = \inf_{\boldsymbol{\eta}} \int q(\boldsymbol{\theta}|\boldsymbol{\eta}) \log \frac{q(\boldsymbol{\theta}|\boldsymbol{\eta})}{p(\mathbf{x}|\boldsymbol{\theta})p(\boldsymbol{\theta})} \, d\boldsymbol{\theta} 
= \inf_{\boldsymbol{\eta}} \mathbb{E}_{q(\boldsymbol{\theta}|\boldsymbol{\eta})}\left[\log q(\boldsymbol{\theta}|\boldsymbol{\eta}) - \log p(\mathbf{x}|\boldsymbol{\theta}) - \log p(\boldsymbol{\theta})\right],
\end{equation*}
where $\boldsymbol{\eta}^*$ represents the optimal parameter estimate for maximizing the ELBO.

We define a loss function to be minimized. Assume that $\varepsilon$ is a random variable with a probability density function $q(\varepsilon)$, and let $f$ be a function. In this case, as shown in \cite{Blundell2015}, if
\begin{equation}\label{e2}
q(\varepsilon) d\varepsilon = q(\boldsymbol{\theta}|\boldsymbol{\eta}) d\boldsymbol{\theta},
\end{equation}
then
\begin{equation}\label{e3}
\frac{\partial}{\partial \boldsymbol{\eta}} \mathbb{E}_{q(\boldsymbol{\theta}|\boldsymbol{\eta})} [f(\boldsymbol{\theta}, \boldsymbol{\eta})] = \mathbb{E}_{q(\varepsilon)} \left[\frac{\partial f(\boldsymbol{\theta}, \boldsymbol{\eta})}{\partial \boldsymbol{\eta}}\right].
\end{equation}

Equation \eqref{e3} shows that if we can find a transformation for $\boldsymbol{\theta}$ such that it depends on $\varepsilon$ and satisfies the condition \eqref{e2}, then samples of $\varepsilon$ can be used to generate samples of $\boldsymbol{\theta}$. This approach focuses on finding $\boldsymbol{\eta}^*$, as defined above. Therefore, we define
\begin{equation}\label{e4}
f(\boldsymbol{\theta}, \boldsymbol{\eta}) = \log q(\boldsymbol{\theta}|\boldsymbol{\eta}) - \log p(\mathbf{x}|\boldsymbol{\theta}) - \log p(\boldsymbol{\theta}).
\end{equation}

Note that the expectation of $f(\boldsymbol{\theta}, \boldsymbol{\eta})$ is the negative of the ELBO and the gradient of $f(\boldsymbol{\theta}, \boldsymbol{\eta})$ provides an unbiased estimate of the gradient defined in equation \eqref{e3}. In addition, increasing the number of samples of $\varepsilon$ reduces the variance of the estimated gradient. This method utilizes equation \eqref{e4} as an {objective function} to determine $\boldsymbol{\eta}^*$.

\section{Variational Bayes Regression Network with Fixed Variance}

{In this section, {we consider regression problems with constant variance.}} 
As we mentioned in section 3, we can use equation \eqref{e4} for the optimization problem as the KL error and we are interested in minimizing it.

Following the approach described in \cite{Blundell2015}, the variance of the likelihood function in a regression problem can be fixed to a constant value. In regression problems, assuming independency among the output features ({i.e., the values of the vector $\mathbf{y}_i$}), we have
{
\[
\mathbf{y}_i = \phi(\mathbf{x}_i; \mathbf{W}) + \varepsilon_i, \quad \varepsilon_i \sim \mathcal{N}(0, \sigma^2_0 \mathbf{I_q}),
\]
where $\mathbf{I_q}$ is the identity matrix, $\mathbf{x}_i \in \mathbb{R}^p$ represents the {$i$-th} input variable, $\mathbf{y}_i \in \mathbb{R}^q$ denotes the {$i$-th} output variable, {$i=1,...,n$}, and $\sigma^2_0$ is a fixed known variance. Moreover, $\phi(\mathbf{x}_i; \mathbf{W)}$ represents the output of a NN with an arbitrary number of layers. In this case, we have the conditional distribution
\[
\mathbf{y}_i | \mathbf{x}_i, \mathbf{W} \sim \mathcal{N}(\phi(\mathbf{x}_i; \mathbf{W}), \sigma^2_0 \mathbf{I_q}).
\]
If $(\mathbf{x}_i, \mathbf{y}_i)$ is the $i$th observation, then the likelihood function, assuming independent observations, is given by
\[
\begin{aligned}
    L(\mathbf{W} \mid \mathbf{y}, \mathbf{x}) &= p(\mathbf{y}_1, \dots, \mathbf{y}_n \mid \mathbf{x}_1, \dots, \mathbf{x}_n, \mathbf{W}) \\
    &= \prod_{i=1}^n \frac{1}{\sqrt{\textcolor{black}{(2 \pi \sigma_0^2)^{q}}}} \exp\left\{-\frac{1}{2 \sigma_0^2} (\mathbf{y}_i - \phi(\mathbf{x}_i; \mathbf{W}))^T (\mathbf{y}_i -\phi(\mathbf{x}_i; \mathbf{W})) \right\}.
\end{aligned}
\]
Thus, \cite{Blundell2015} considered the objective function 
\begin{equation}\label{e42}
f(\mathbf{W}, \boldsymbol{\eta}) = \log q(\mathbf{W}|\boldsymbol{\eta}) - \log L(\mathbf{W} \mid \mathbf{y}, \mathbf{x}) - \log p(\mathbf{W}),
\end{equation}
\textcolor{black}{where $q(\mathbf{W}|\boldsymbol{\eta}) = \mathcal{N}(\mathbf{W}; \boldsymbol{\mu}_w, \operatorname{diag}(\boldsymbol{\sigma}_w^2))$ with $\boldsymbol{\sigma}_w = \log(1+\exp(\boldsymbol{\rho}_w))$, and $\boldsymbol{\eta} = (\boldsymbol{\mu}_w, {\boldsymbol{\rho}_w})$, seeking the optimum value of $\boldsymbol{\eta}$}. We denote this method by VBNET-FIXED throughout the remaining of the paper. 
}
However, this approach fails to effectively capture uncertainty in the variance parameter. In the {forthcoming} section, we will discuss how to introduce uncertainty in the variance by using  parametrization, so that the model can be generalized by considering the posterior distribution over the variance parameter.

\section{Variational Bayes Regression Network with Variance Uncertainty}

In this section, we suggest to use a variational posterior distribution over weights, biases, and also the variance of the likelihood function. {{Our proposal} can be considered as a generalization of the previous model introduced in section 4}. In our approach, we assume that the posterior parameters are represented by \( \boldsymbol{\theta} = (\mathbf{W}, S) \), where \( \mathbf{W} \) encompasses the weights and biases of the NN, and \( S \) denotes a parameter related to the variance of the likelihood function. Additionally, we define the variational parameters as \( \boldsymbol{\eta} = (\boldsymbol{\sigma}_w^2, \boldsymbol{\mu}_w, \sigma_L^2, \mu_L) \), where \( \boldsymbol{\sigma}_w^2 \) and \( \boldsymbol{\mu}_w \) pertain to the variational distribution over \( \mathbf{W} \), and \( \sigma_L^2 \) and \( \mu_L \) are associated to variational distribution over \( S \).

{
Recall in a regression task we have
\[
\mathbf{y}_i=\phi\left(\mathbf{x}_i;W\right)+\varepsilon_i={\hat{\mathbf{y}}}_i+\varepsilon_i,\ \ \ \ \ \ \varepsilon_i \sim N(0,\ \textcolor{black}{g(S)} \mathbf{I_q}),
\]
\textcolor{black}{where \( g: \mathbb{R} \rightarrow (0, \infty) \) is a strictly positive transformation function mapping the unconstrained posterior parameter \( S \) to a valid variance value. In this work, we employ the softplus function, \( g(S) = \log(1 + \exp(S)) \), to guarantee the positivity of the variance parameter. Consequently, the conditional distribution of \( \mathbf{y}_i \) is given by:}
\[
\mathbf{y}_i|\mathbf{x}_i \sim N({\hat{y}}_i,\textcolor{black}{g(S)} \mathbf{I_q}), \quad \textcolor{black}{g(S) = \log (1+ \exp \{S\})}.
\]
Therefore, the likelihood function has form
{
\[
\begin{aligned}
    L(\mathbf{W},S|\mathbf{x},\mathbf{y}) &= \prod_{i=1}^n \frac{1}{\sqrt{(2 \pi)^q (\textcolor{black}{g(S)})^q}} \exp\left\{-\frac{1}{2 \textcolor{black}{g(S)}} (\mathbf{y}_i - \phi(\mathbf{x}_i; \mathbf{W}))^T (\mathbf{y}_i -\phi(\mathbf{x}_i; \mathbf{W})) \right\}.
\end{aligned}
\]
}}
Based on \cite{Blundell2015} \textcolor{black}{and utilizing the mean-field approximation \(q(\mathbf{W}, S) = q(\mathbf{W})q(S)\)}, we assume that the variational posterior of all parameters is a diagonal Gaussian distribution, i.e., \textcolor{black}{$\mathbf{W}| \boldsymbol{\mu}_w, \boldsymbol{\sigma}_w^2 \sim N(\boldsymbol{\mu}_w, \operatorname{diag}(\boldsymbol{\sigma}_w^2))$}, where \( |\mathbf{W}| \) denotes the number of weights and biases in the NN, and $\boldsymbol{\sigma}_w = \log \left(1 + \exp(\boldsymbol{\rho}_w)\right)>0$. Similarly, for the posterior parameter \(S\), we assume 
\[
{S|\mu_L,\sigma_L^2 \sim N(\mu_L, \sigma_L^2)},
\]
\textcolor{black}{where its standard deviation is parameterized as}
\[
\textcolor{black}{\sigma_L = \log \left(1+\exp(\rho_L)\right).}
\]
Further, \textcolor{black}{by assuming} \( \boldsymbol{\varepsilon}_w \sim \mathcal{N}(0, I_{|\mathbf{W}|}) \) and \( \varepsilon_L \sim \mathcal{N}(0, 1) \), and applying the reparameterization trick, the posterior samples of \( \mathbf{W} \) and \( S \) are given by
\[\mathbf{W} = \boldsymbol{\mu}_w + \boldsymbol{\varepsilon}_w \odot \textcolor{black}{\boldsymbol{\sigma}_w},
\]
\[S = \mu_L + \varepsilon_L \textcolor{black}{\sigma_L},
\]
where \( \odot \) denotes element-wise multiplication. Here, \( \boldsymbol{\mu}_w \) and \( \boldsymbol{\rho}_w \) are the variational parameters associated with \( \mathbf{W} \), and \( \mu_L \) and \( \rho_L \) are the variational parameters associated with \( S \).

\textcolor{black}{Since $\boldsymbol{\sigma}_w$ and $\sigma_L$ are parameterized by $\boldsymbol{\rho}_w$ and $\rho_L$, respectively, the set of variational parameters to be optimized can be written as $\boldsymbol{\eta} = (\boldsymbol{\rho}_w, \boldsymbol{\mu}_w, \rho_L, \mu_L)$.}

\textcolor{black}{\subsection{Objective function and gradient estimation}}
\textcolor{black}{
In order to minimize the negative $\rm ELBO$, with respect to the variational parameters \(\boldsymbol{\eta} = (\boldsymbol{\rho}_w, \boldsymbol{\mu}_w, \rho_L, \mu_L)\), which involves intractable expectations, we employ Stochastic Gradient Descent (SGD). First, recall that the objective function (negative ELBO) for our proposed method can be written as:
\[
\mathcal{F}(\boldsymbol{\eta}) = \mathbb{E}_{q(\mathbf{W},S|\boldsymbol{\eta})} [f(\mathbf{W},S,\boldsymbol{\eta})],
\]
where \(q(\mathbf{W},S|\boldsymbol{\eta}) = \mathcal{N}(\mathbf{W};\boldsymbol{\mu}_w, \operatorname{diag}(\boldsymbol{\sigma}^2_w)) \mathcal{N}(S;\mu_L,\sigma^2_L)\) is a diagonal Gaussian variational posterior over \(\mathbf{W}\) and \(S\), \(p(\mathbf{W},S)=p(\mathbf{W})p(S)\) represents the factorized prior (typically Gaussian, acting as a regularizer), and \(f(\mathbf{W},S,\boldsymbol{\eta})\) is the cost function defined as:
\[
f(\mathbf{W},S,\boldsymbol{\eta}) = \log q(\mathbf{W}|\boldsymbol{\mu}_w, \boldsymbol{\sigma}^2_w) + \log q(S|\mu_L,\sigma^2_L) - \log p(\mathbf{W}) - \log p(S) - \log L(\mathbf{W},S|\mathbf{x},\mathbf{y}).
\]
}
\textcolor{black}{
	It is important to note that the KL divergence term is implicitly minimized via the term $\mathbb{E}_{q}[f(\mathbf{W},S,\boldsymbol{\eta})]$. We use a Monte Carlo approximation for this term rather than an analytic closed-form solution. This approach avoids the need for tedious derivation of the KL divergence for complex priors and allows for greater flexibility in the choice of prior distributions.
}

\textcolor{black}{
To minimize this objective, we compute the gradients of \(\mathcal{F}(\boldsymbol{\eta})\) with respect to \(\boldsymbol{\eta}\). Using standard properties of expectations (via the Dominated Convergence Theorem) and applying the reparameterization trick, the gradients can be expressed as expectations over the variable \(\varepsilon\):
\[
\nabla_{\boldsymbol{\mu}_w} \mathcal{F}(\boldsymbol{\eta}) = \mathbb{E}_{\boldsymbol{\varepsilon}_w \sim \mathcal{N}(0,I_{|\mathbf{W}|})} \left[\frac{\partial f}{\partial \mathbf{W}} + \frac{\partial f}{\partial \boldsymbol{\mu}_w}\right],
\]
\[
\nabla_{\boldsymbol{\rho}_w} \mathcal{F}(\boldsymbol{\eta}) = \mathbb{E}_{\boldsymbol{\varepsilon}_w \sim \mathcal{N}(0, I_{|\mathbf{W}|})} \left[\frac{\partial f}{\partial \mathbf{W}} \odot \frac{\boldsymbol{\varepsilon}_w}{1 + \exp \{-\boldsymbol{\rho}_w\}} + \frac{\partial f}{\partial \boldsymbol{\rho}_w}\right].
\]
Similarly, for the variational parameters \(\mu_L\) and \(\rho_L\), we have:
\[
\nabla_{\mu_L} \mathcal{F}(\boldsymbol{\eta}) = \mathbb{E}_{\varepsilon_L \sim \mathcal{N}(0,1)} \left[\frac{\partial f}{\partial S} + \frac{\partial f}{\partial \mu_L}\right],
\]
\[
\nabla_{\rho_L} \mathcal{F}(\boldsymbol{\eta}) = \mathbb{E}_{\varepsilon_L \sim \mathcal{N} (0,1)} \left[\frac{\partial f}{\partial S} \cdot \frac{\varepsilon_L}{ 1 + \exp \{-\rho_L\}} + \frac{\partial f}{\partial \rho_L}\right].
\]
Since the expectations above are analytically intractable, we use a Monte Carlo estimator. Let \(\boldsymbol{\varepsilon}^1_w, ...,\boldsymbol{\varepsilon}^M_w\) and \(\varepsilon^1_L, ..., \varepsilon^M_L\) be i.i.d. samples from \(\mathcal{N}(0,I_{|\mathbf{W}|})\) and \(\mathcal{N}(0,1)\), respectively. By applying the reparameterization trick, let \(\mathbf{W}_j = \boldsymbol{\mu}_w + \boldsymbol{\varepsilon}^j_w \odot \boldsymbol{\sigma}_w\) and \(S_j = \mu_L + \varepsilon^j_L \sigma_L\) denote the posterior samples for \(j=1,...,M\). The gradient estimator for \(\boldsymbol{\mu}_w\) is defined as:
\begin{equation} \label{gradient_estimator}
	\widehat{\nabla}_{\boldsymbol{\mu}_w} = \frac{1}{M} \sum_{j=1}^M \left[\frac{\partial f(\mathbf{W},S,\boldsymbol{\eta})}{\partial \mathbf{W}} + \frac{\partial f(\mathbf{W},S,\boldsymbol{\eta})}{\partial \boldsymbol{\mu}_w}\right]_{\mathbf{W}=\mathbf{W}_j, S=S_j},
\end{equation}
which is an unbiased estimator. The estimators for \(\boldsymbol{\rho}_w, \mu_L\), and \(\rho_L\) are defined analogously.
}

Algorithm \ref{al1} describes the VB algorithm for the regression network with variance uncertainty. { We denote this method by VBNET-SVAR throughout the remaining of the paper.}

\textcolor{black}{
	Note that the computational complexity of VBNET-SVAR is comparable to VBNET-FIXED, as only two additional scalar parameters have been introduced and is independent of size of the neural network.
}

\begin{algorithm}
\caption{Variational Inference Algorithm for regression network with variance uncertainty}\label{al1}
\begin{algorithmic}
	\STATE \textbf{1.} \textcolor{black}{Initialize the variational parameters \(\boldsymbol{\eta}=(\boldsymbol{\mu}_w, \boldsymbol{\rho}_w, \mu_L, \rho_L)\) randomly.}
    \STATE \textbf{2.} \textcolor{black}{Sample $\boldsymbol{\varepsilon}^1_w,...,\boldsymbol{\varepsilon}^M_w \overset{\text{iid}}{\sim} \mathcal{N}(0, I_{|\mathbf{W}|})$ and $\varepsilon^1_L,...,\varepsilon^M_L \overset{\text{iid}}{\sim} \mathcal{N}(0, 1)$}.
    
    \STATE \textbf{3.} Compute posterior samples:
    \[
    \textcolor{black}{\mathbf{W}_j = \boldsymbol{\mu}_w + \boldsymbol{\varepsilon}^j_w \odot \log (1 + \exp(\boldsymbol{\rho}_w)), \quad \textcolor{black}{S_j = \mu_L + \varepsilon^j_L \log(1 + \exp(\rho_L)), \quad j=1,...,M.}}
    \]
 
    \STATE \textbf{4.} Define the objective function:
    \[
    \textcolor{black}{f(\mathbf{W}, S, \boldsymbol{\eta}) = \log q(\mathbf{W}|\boldsymbol{\mu}_w, \boldsymbol{\sigma}_w^2) + \log q(S|\mu_L, \sigma_L^2) - \log p(\mathbf{W}) - \log p(S) -  \log L(\mathbf{W},S|\mathbf{x},\mathbf{y}).}
    \]
    
    \STATE \textbf{5.} \textcolor{black}{Calculate the gradients using equation \eqref{gradient_estimator}}:
    \[
    \textcolor{black}{\widehat{\nabla}_{\boldsymbol{\mu}_w} = \frac{1}{M} \sum_{j=1}^M \left[\frac{\partial f}{\partial \mathbf{W}} + \frac{\partial f}{\partial \boldsymbol{\mu}_w}\right]_{\mathbf{W}=\mathbf{W}_j, S=S_j},}
    \]
    \[
    \textcolor{black}{\widehat{\nabla}_{\boldsymbol{\rho}_w} = \frac{1}{M} \sum_{j=1}^M \left[\frac{\partial f}{\partial \mathbf{W}}\odot \frac{\boldsymbol{\varepsilon}^j_w}{1  + \exp \{-\boldsymbol{\rho}_w\}} + \frac{\partial f}{\partial \boldsymbol{\rho}_w}\right]_{\mathbf{W}=\mathbf{W}_j, S=S_j},}
    \]
    \[
    \textcolor{black}{\widehat{\nabla}_{\mu_L} = \frac{1}{M} \sum_{j=1}^M \left[\frac{\partial f}{\partial S} + \frac{\partial f}{\partial \mu_L}\right]_{\textbf{W}=\textbf{W}_j, S=S_j},}
    \]
    \[
    \textcolor{black}{\widehat{\nabla}_{\rho_L} = \frac{1}{M} \sum_{j=1}^M \left[\frac{\partial f}{\partial S} \cdot \frac{\varepsilon^j_L}{1  + \exp \{-\rho_L\}} + \frac{\partial f}{\partial \rho_L}\right]_{\textbf{W}=\textbf{W}_j, S=S_j}.}
    \]
    
    \STATE \textbf{6.} Update the variational parameters:
    \[
    \textcolor{black}{\boldsymbol{\mu}_w \leftarrow \boldsymbol{\mu}_w - \gamma_w \, \widehat{\nabla}_{\boldsymbol{\mu}_w}, \quad   \boldsymbol{\rho}_w \leftarrow \boldsymbol{\rho}_w - \gamma_w \, \widehat{\nabla}_{\boldsymbol{\rho}_w},}
    \]
    \[
    \textcolor{black}{\mu_L \leftarrow \mu_L - \gamma_l \, \widehat{\nabla}_{\mu_L}, \quad  \rho_L \leftarrow \rho_L - \gamma_l \, \widehat{\nabla}_{\rho_L}.}
    \]
    
    \STATE \textbf{7.} \textcolor{black}{Repeat from step 2 until convergence.}
\end{algorithmic}
\end{algorithm}

\section{Experimental Results}
In this section, we evaluate the results of our proposed model {on two regression problems under the following scenarios}
{
\begin{itemize}
    \item The normal prior, which corresponds to a normal distribution, assuming independency between the prior over parameters, is given by
\begin{equation*}
p(\mathbf{W}) = \frac{1}{\sqrt{(2 \pi \sigma_p^2)^{|\mathbf{W}|}}} \exp\left(-\frac{1}{2 \sigma_p^2} vec(\mathbf{W})^\top vec(\mathbf{W})\right),
\end{equation*}
where $\sigma_p^2$ denotes the variance of the prior, and $\textbf{W}$ represents the vector of parameters.
\item The spike-and-slab prior \cite{ssp}, which corresponds to a mixture of normal distributions, assuming independency between the prior over parameters, is given by
\[p(\mathbf{W}) = \prod_j \left[ z_j \mathcal{N}(W_j; 0, \sigma_1^2) + (1 - z_j) \mathcal{N}(W_j; 0, \sigma_2^2) \right],
\]
where, $W_j$ represents the $j$th (weight or bias) parameter in the model, $\sigma_1^2$ denotes the variance of the first mixture component, which is greater than $\sigma_2^2$, the variance of the second mixture component, and 
$$z_j \stackrel{\rm \tiny indep.}{\sim}{\rm Ber}(\pi).$$
The second variance $\sigma_2^2$ is assumed to be small ($\sigma_2^2 \approx 0$), allowing the prior to concentrate samples around 0 with probability $1 - \pi$.
\end{itemize}
}
{We compare the performance of our proposed model} with various models, including the model introduced in \cite{Blundell2015}, and the frequentist NN. The results indicate that variance uncertainty can significantly enhance performance. It should be noted that in the nonlinear function estimation study, we have set the fixed variance for the model proposed by \cite{Blundell2015} to the train set MSE of the frequentist NN, while in the riboflavin data analysis, the fixed variance is set to the maximum of $0.2 {\rm var}(y_{\rm train})$ and the train set MSE of the frequentist NN to prevent the overfit small error problem caused by the frequentist NN. These values are chosen by trial and error to obtain the best performance of the model of \cite{Blundell2015} (instead of a cross-validation approach).

\subsection{Nonlinear function estimation}
Assume that {in the problem of regression curve,} the { target variable is} generated according to the model of the following curve.

{ \[y = x + 2 \sin \left( 2 \pi ( x + \varepsilon ) \right) + 2 \sin \left( 4 \pi ( x + \varepsilon ) \right) + \varepsilon,
\]
where $\varepsilon \sim N(0,0.02)$.}
{ To model the out-of-sample unforeseen risks, the values of $x$ for the train and test samples are generated from uniform distributions with supports  $[-0.1, 0.6]$ and $[-0.25, 0.85]$, respectively.} { The train and test sets contain 800 and 200 samples, respectively.}

\begin{figure}
\centering
\centerline{\includegraphics[scale=0.7]{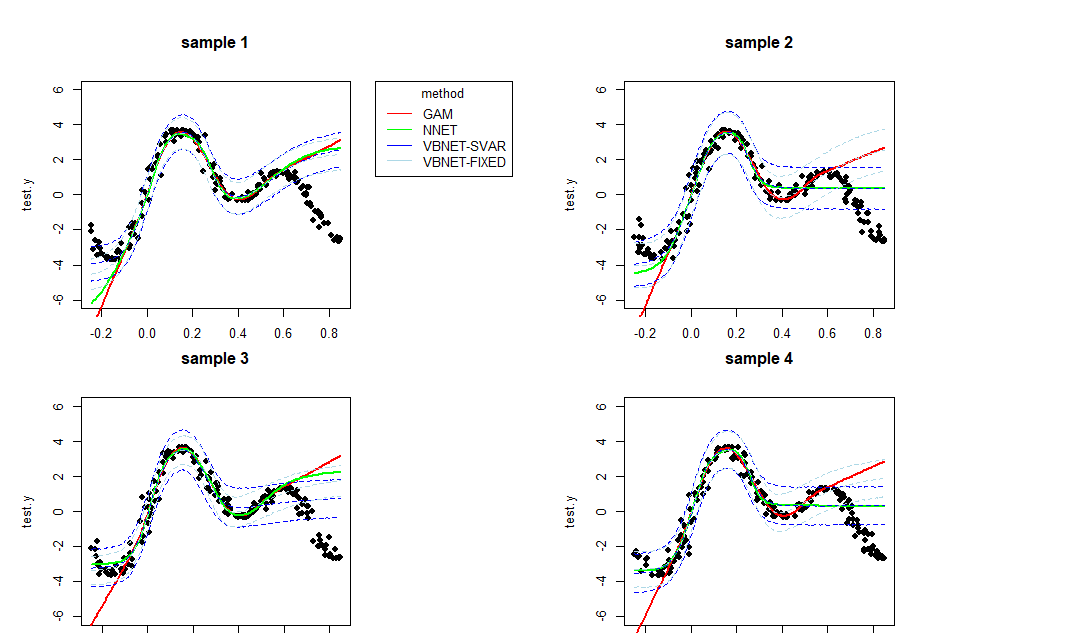}}
\caption{The result of samples 1-4 for the regression curve simulation study.}\label{fig1}
\end{figure}

\begin{figure}
\centering
\centerline{\includegraphics[scale=0.5]{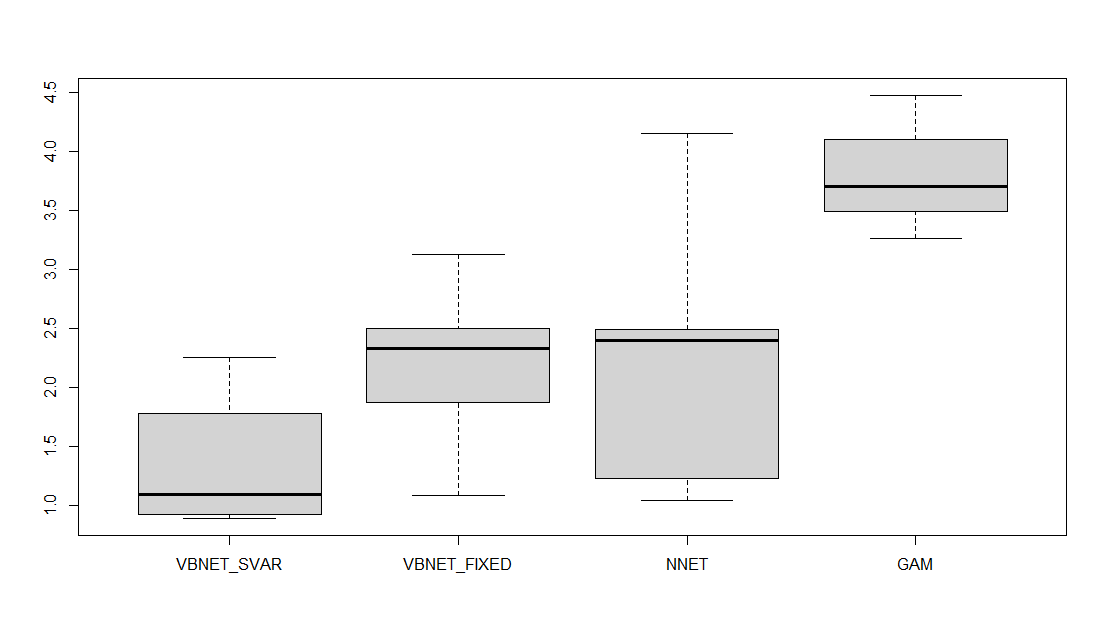}}
\caption{ The box plots of the MSPE for the regression curve simulation study.}\label{fig2}
\end{figure}

\begin{figure}
\centering
\centerline{\includegraphics[scale=0.5]{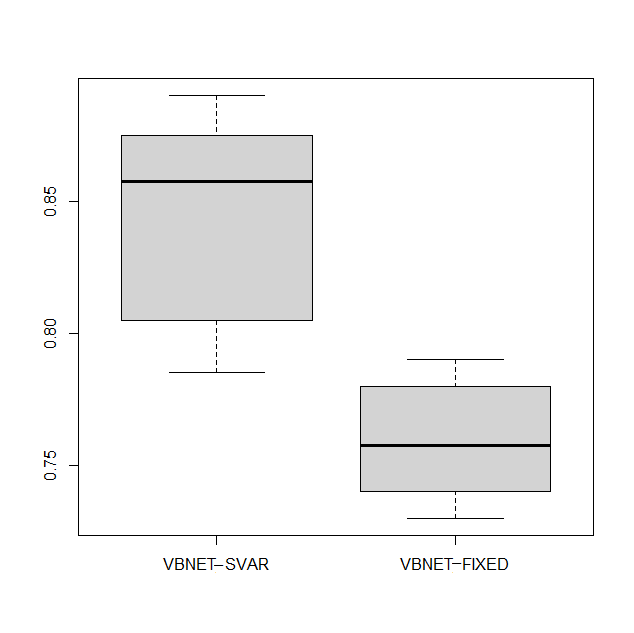}}
\caption{ The box plots of the coverage probabilities for the regression curve simulation study.}\label{fig3}
\end{figure}

{  The prior \( p(\mathbf{W}) = \mathcal{N}(\mathbf{W}; 0, \sigma_p^2 I_{|\mathbf{W}|}) \) is used for the VBNET-FIXED, while and the prior \( p(\mathbf{W}, S) = \mathcal{N}(\mathbf{W}; 0, \sigma_{pw}^2 I_{|\mathbf{W}|}) \mathcal{N}(S; 0, \sigma_{ps}^2) \) is considered for the VBNET-SVAR. Four competitive models are considered for the comparison study; VBNET-SVAR, VBNET-FIXED, the standard NN (NNET), and the Generalized Additive Model (GAM \cite{gam}). A total number of 10 replications is performed in this simulation study.}

{ Figure \ref{fig1} shows the estimated curves for all competitive models, in which the two Bayesian models also include 95\% prediction intervals. Figure \ref{fig2} presents box plots for the values of the Mean Squared Prediction Error (MSPE) for each model, defined as follows
$$ {\rm MSPE} = \frac{1}{m}\sum_{i=1}^{m}(y_i-\hat{y}_i)^2,$$
where $y_i,\; i=1,\ldots,m$ are the target values of the test set, and $\hat{y}_i$ is the predicted value for $y_i$. From Figure \ref{fig2}, one can observe that the VBNET-SVAR model performs better than the other competitive models for out-of-sample prediction. Figure \ref{fig3} shows the box plot for the coverage probabilities of the two Bayesian models for the test set, demonstrating that the VBNET-SVAR model provides more data coverage for the out-of-sample prediction purpose.}

\subsection{Riboflavin dataset: PCA-BNN Scenario}
Riboflavin is a dataset with a small number of samples but many features. This makes it a suitable dataset for Bayesian models.
Dataset of riboflavin production by Bacillus subtilis \cite{ribo} contains n = 71 observations of p = 4088 predictors (gene expressions) and a one-dimensional response (riboflavin production). From 71 samples, 56 randomly drawn samples are considered as the train set, and the remaining 15 samples are considered as the test set. We have repeated this sampling 10 times to evaluate the performance of the competitor models on this data set.

\textcolor{black}{
In this section, we examine the {PCA-BNN} scenario. To address the curse of dimensionality, we applied Principal Component Analysis (PCA) and selected the first 10 principal components as the input features for all models. The neural network architecture consisted of two hidden layers with 10 and 15 neurons, respectively, employing sigmoid activation functions.
For the Bayesian models, we assigned independent Gaussian priors over the weights and the variance parameter as $p(\mathbf{W}, S) = \mathcal{N}(\mathbf{W}; \mathbf{0}, 100\mathbf{I})\mathcal{N}(S; 0, 100)$. We compared our proposed method ({VBNET-SVAR}) against the fixed-variance BNN ({VBNET-FIXED}), a Standard Neural Network (NNET), and a Generalized Additive Model (GAM).
}

\textcolor{black}{
\textbf{Results and Discussion:}
Table \ref{tab:ribo_pca_results} summarizes the quantitative results averaged over 10 independent runs. We report the Mean Squared Prediction Error (MSPE), Average Prediction Interval Width (PI Width), and Coverage Probability (CP), along with their Standard Errors (SE).
}

\begin{table}[ht]
	\color{black}
	\captionsetup{labelfont={color=black}, textfont={color=black}}
	\centering
	\resizebox{\textwidth}{!}{%
		\begin{tabular}{lccc}
			\hline
			\textbf{Model} & \textbf{MSPE (SE)} & \textbf{Avg PI Width (SE)} & \textbf{Coverage Prob. (SE)} \\ \hline
			\textbf{VBNET-SVAR} & \textbf{0.7891 (0.0814)} & 4.5022 (0.0699) & \textbf{0.9800 (0.0102)} \\
			VBNET-FIXED & 1.4006 (0.1784) & 2.8242 (0.0411) & 0.8000 (0.0298) \\ \hline
			GAM & 1.3396 (0.1839) & - & - \\
			NNET & 0.9177 (0.1274) & - & - \\ \hline
		\end{tabular}%
		
	}
	\caption{Performance comparison on the Riboflavin dataset (PCA-BNN scenario) over 10 independent runs. Values represent Mean (Standard Error).}
	\label{tab:ribo_pca_results}
\end{table}

\begin{figure}
	\captionsetup{labelfont={color=black}, textfont={color=black}}
	\centering
	\includegraphics[width=1\textwidth]{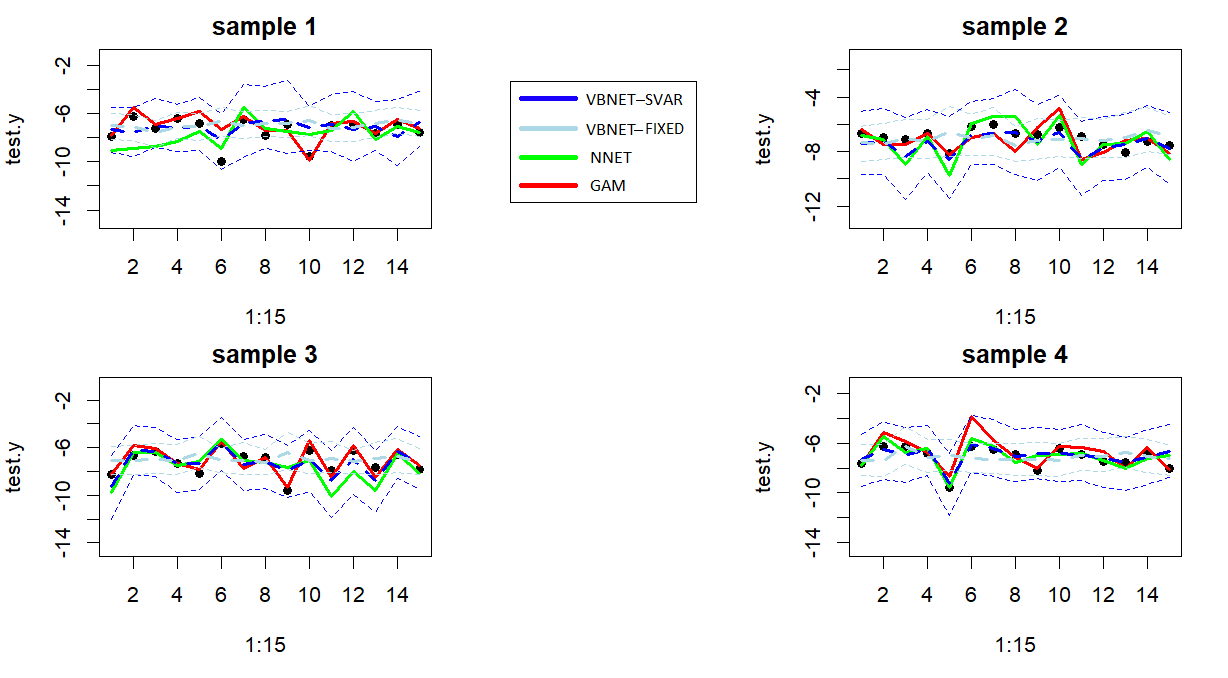}
	\caption{Prediction results for samples 1-4 on the Riboflavin dataset (PCA-BNN scenario). The blue dashed lines represent the 95\% prediction intervals of the proposed VBNET-SVAR model, which successfully cover the target values.}\label{fig:ribo_pca_pred}
\end{figure}

\textcolor{black}{
As shown in Table \ref{tab:ribo_pca_results}, {VBNET-SVAR} achieves the lowest prediction error (MSPE = 0.7891), significantly outperforming VBNET-FIXED (MSPE = 1.4006) and the frequentist baselines.
Figure \ref{fig:ribo_pca_pred} illustrates the predicted values and 95\% prediction intervals for four random test samples. The results highlight a key advantage of the proposed method:
}
\begin{itemize}
	\item \textcolor{black}{
	\textbf{Uncertainty Quantification:} VBNET-FIXED produces overly narrow prediction intervals (Avg Width $\approx$ 2.82), leading to a low coverage probability of 0.80. This suggests the model is overconfident and underestimates the noise introduced by dimensionality reduction.}
	\item \textcolor{black}{
	\textbf{Robustness:} In contrast, VBNET-SVAR learns the posterior distribution of the variance, adapting the intervals (Avg Width $\approx$ 4.50) to fully encompass the epistemic uncertainty. This results in a coverage probability of 0.98, demonstrating that the model provides safer and more reliable predictions in this high-uncertainty regime.
	}
\end{itemize}

\begin{figure}
	\captionsetup{labelfont={color=black}, textfont={color=black}}
	\centering
	\includegraphics[width=1.0\textwidth]{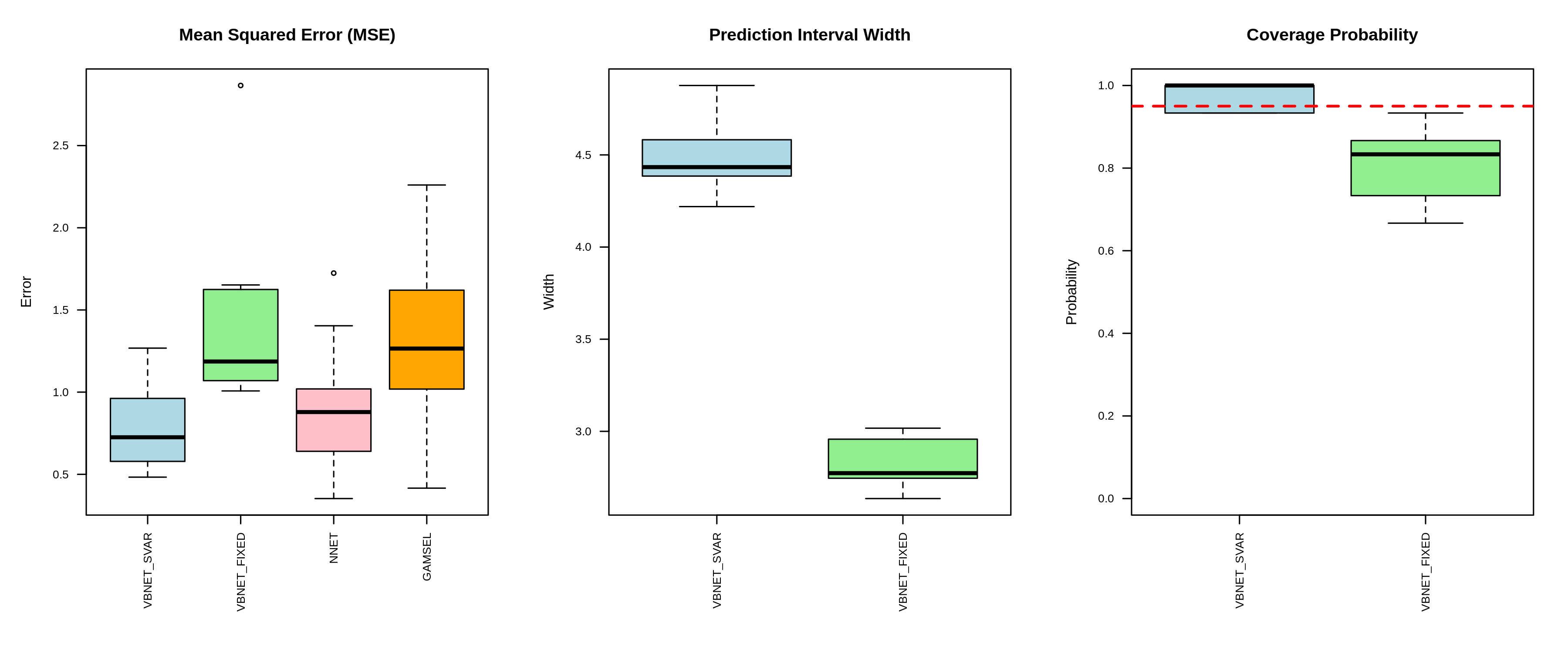}
	\caption{Performance comparison for the PCA-BNN scenario over 10 independent runs. Left: MSPE (lower is better); Middle: Prediction Interval Width; Right: Coverage Probability (dashed line represents the nominal 95\% level).}\label{fig:ribo_pca_boxplot}
\end{figure}

\textcolor{black}{Figure \ref{fig:ribo_pca_boxplot} further confirms these findings, showing the distribution of MSPE, interval width, and coverage probability across all 10 runs.}

\begin{figure}
	\captionsetup{labelfont={color=black}, textfont={color=black}}
	\centering
	\includegraphics[width=0.8\textwidth]{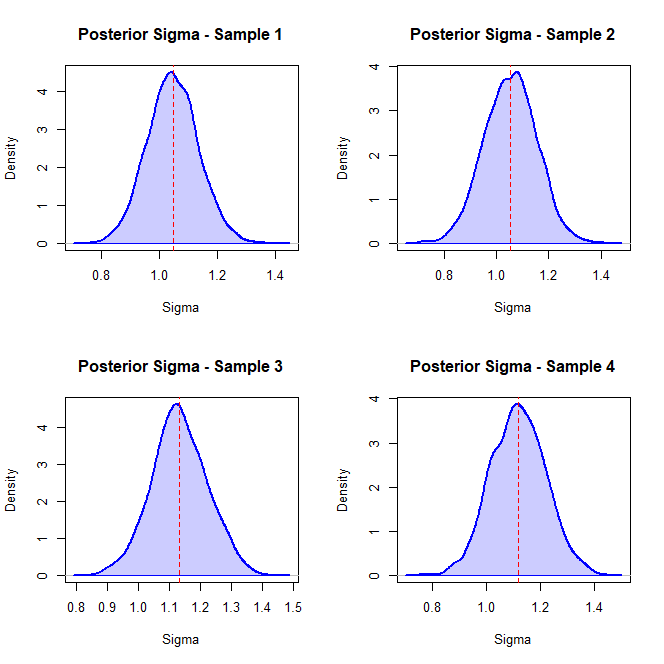}
	\caption{The approximate Posterior distribution of the likelihood variance parameter ($\sqrt{g(S)}$) for different test samples in the PCA-BNN scenario.}\label{fig:ribo_pca_sigma}
\end{figure}

\textcolor{black}{Figure \ref{fig:ribo_pca_sigma} illustrates the approximate posterior distribution of the variance parameter for different test samples. By integrating over this distribution during prediction, VBNET-SVAR yields the robust and well-calibrated intervals observed in the results.}

\subsection{Riboflavin dataset: Dropout-BNN Scenario}
\textcolor{black}{In the second scenario, we utilized the full feature space ($p=4088$) without dimensionality reduction. To handle the high dimensionality, we employed a Bayesian Dropout approach imposed by a spike-and-slab prior. We compared {VBNET-SVAR} against {VBNET-FIXED}, a Neural Network with Dropout (DropoutNN), and the sparse-GAM model (GAMSEL) \cite{gamsel}. The network architecture remained two hidden layers with 128 and 64 neurons, and results were averaged over 10 independent runs.}

\textcolor{black}{\textbf{Results and Discussion:}
	Table \ref{tab:ribo_dropout_results} summarizes the performance metrics. Consistent with the PCA scenario, our proposed method demonstrates superior performance.}

\begin{table}[ht]
	\color{black}
	\captionsetup{labelfont={color=black}, textfont={color=black}}
	\centering
	\resizebox{\textwidth}{!}{%
		\begin{tabular}{lccc}
			\hline
			\textbf{Model} & \textbf{MSPE (SE)} & \textbf{Avg PI Width (SE)} & \textbf{Coverage Prob. (SE)} \\ \hline
			\textbf{VBNET-SVAR} & \textbf{0.3077 (0.0614)} & 3.8370 (0.0378) & \textbf{1.0000 (0.0000)} \\
			VBNET-FIXED & 0.3607 (0.0603) & 1.3412 (0.0216) & 0.7200 (0.0490) \\ \hline
			GAMSEL & 0.4537 (0.0977) & - & - \\
			NN-Dropout & 0.4893 (0.1150) & - & - \\ \hline
		\end{tabular}%
	}
	\caption{Performance comparison on the Riboflavin dataset (Dropout-BNN scenario) over 10 independent runs. Values represent Mean (Standard Error).}
	\label{tab:ribo_dropout_results}
\end{table}

\textcolor{black}{The results in Table \ref{tab:ribo_dropout_results} demonstrate the significant advantage of modeling variance uncertainty in high-dimensional settings:}
\begin{enumerate}
	\item \textcolor{black}{\textbf{Prediction Accuracy:} VBNET-SVAR achieves the best predictive performance with an MSPE of 0.3077, outperforming both the fixed-variance BNN (0.3607) and the frequentist baselines (GAMSEL and NN-Dropout).}
	\item \textcolor{black}{\textbf{Uncertainty and Robustness:} A striking difference is observed in uncertainty quantification. VBNET-FIXED exhibits a coverage probability of only 0.72, which is far below the nominal 95\% level. This indicates severe \textit{overconfidence}, where the model underestimates the noise (Avg Width $\approx$ 1.34) and fails to capture the true targets. In contrast, VBNET-SVAR adapts to the high epistemic uncertainty inherent in this $p \gg n$ regime. By learning a wider variance, it produces conservative prediction intervals (Avg Width $\approx$ 3.84) that achieve 100\% coverage, ensuring that the model is robust and reliable for decision-making.}
\end{enumerate}

\begin{figure}
	\captionsetup{labelfont={color=black}, textfont={color=black}}
	\centering
	\includegraphics[width=0.89\textwidth]{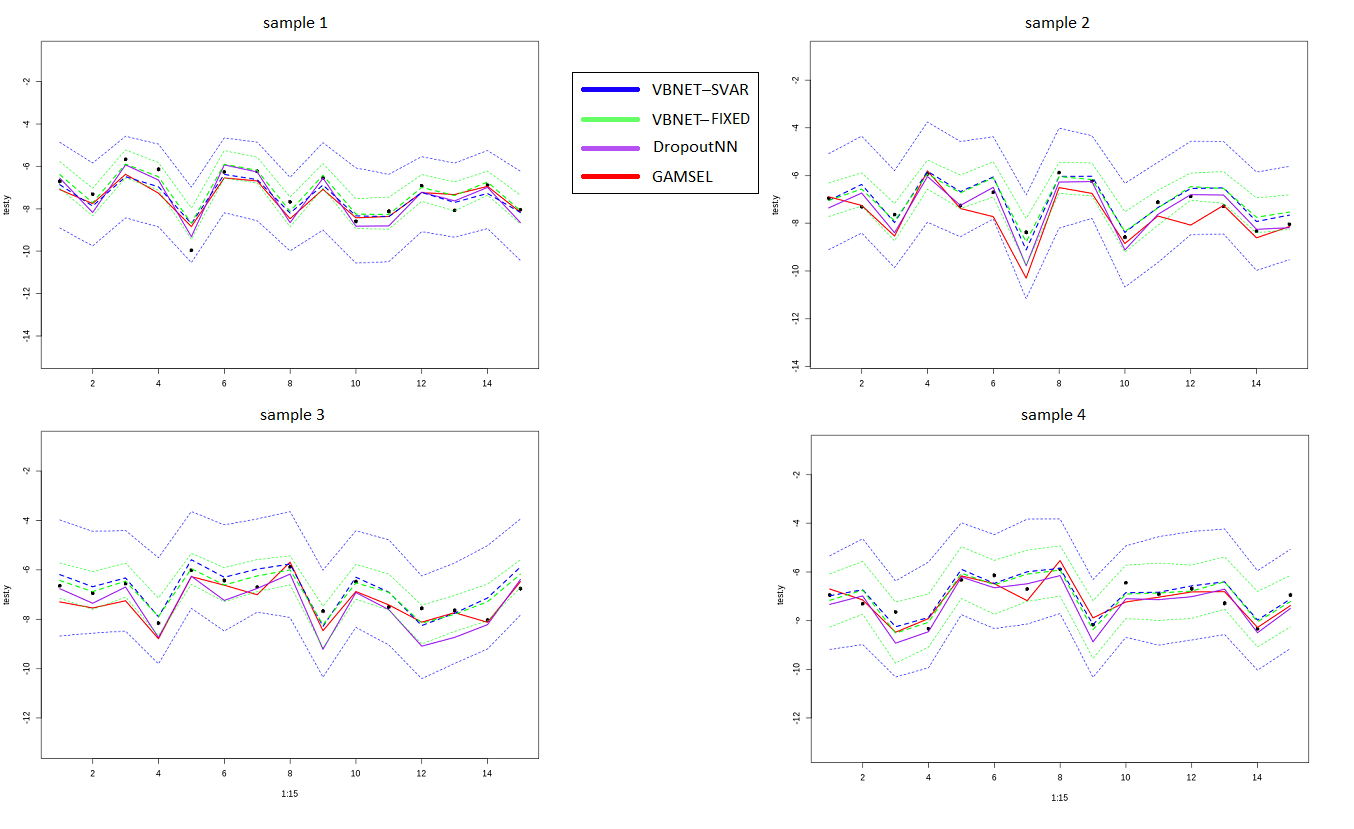}
	\caption{Prediction results for samples 1-4 on the Riboflavin dataset (Dropout-BNN scenario). The blue dashed lines (VBNET-SVAR) provide robust 95\% prediction intervals that fully cover the data, while green dashed lines (VBNET-FIXED) are overly narrow.}\label{fig:ribo_dropout_pred}
\end{figure}

\begin{figure}
	\captionsetup{labelfont={color=black}, textfont={color=black}}
	\centering
	\includegraphics[width=1.0\textwidth]{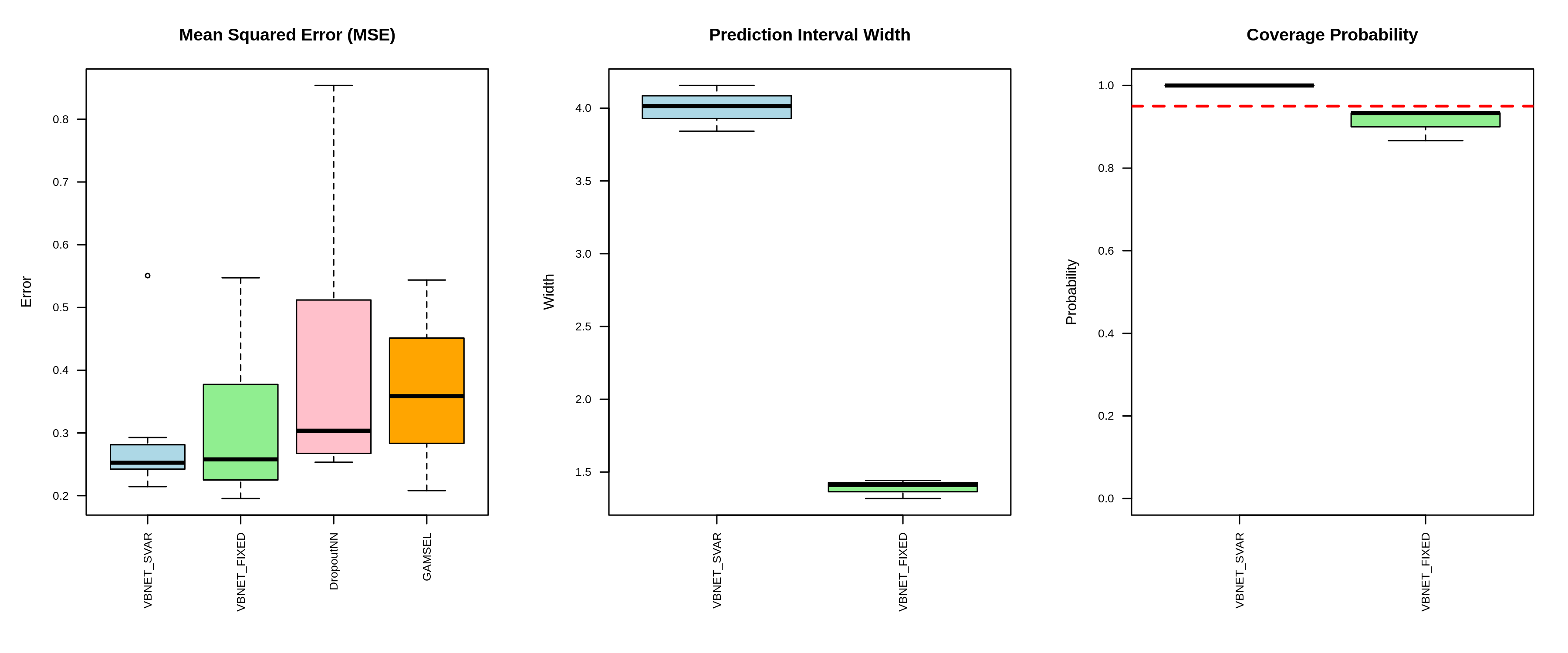}
	\caption{Performance metrics for the Dropout-BNN scenario over 10 independent runs. Left: MSPE; Middle: PI Width; Right: Coverage Probability. The red dashed line represents the 95\% nominal coverage.}\label{fig:ribo_dropout_boxplot}
\end{figure}

\begin{figure}
	\captionsetup{labelfont={color=black}, textfont={color=black}}
	\centering
	\includegraphics[width=0.8\textwidth]{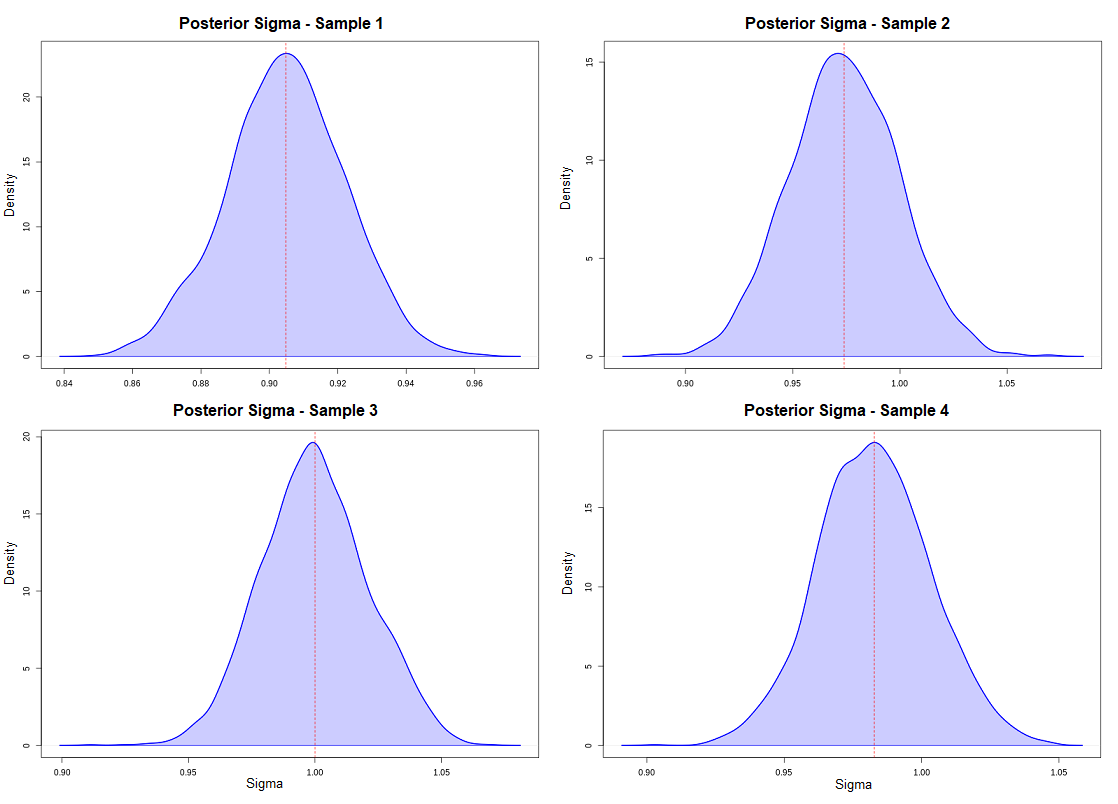}
	\caption{The approximate Posterior distribution of the likelihood variance parameter ($\sqrt{g(S)}$) for different test samples in the Dropout-BNN scenario.}\label{fig:ribo_dropout_sigma}
\end{figure}

\textcolor{black}{Figure \ref{fig:ribo_dropout_pred} visualizes these predictions for four random samples. 
	Figure \ref{fig:ribo_dropout_boxplot} compares the distribution of metrics across runs. The boxplots confirm that VBNET-SVAR consistently maintains lower error and superior coverage compared to the baselines.
	Additionally, Figure \ref{fig:ribo_dropout_sigma} displays the posterior density of the variance parameter ($\sqrt{g(S)}$) learned by VBNET-SVAR, confirming that the method successfully infers the uncertainty associated with the observation noise.}

\section{Concluding remarks}

In this paper, we {considered} variance uncertainty in neural networks, specialized for the regression tasks. Based on the MSPE and coverage probabilities, {a simulation study of a simple function approximation problem and a real genetic data set analysis showed that the prediction performance of the BNN improved by considering variance uncertainty}. This method applies when no information is available regarding the variance setting of the likelihood function, which is common in many real-world applications. This model can also generalize BNNs in regression problems {with fixed variance}. The code for this study is available { online at github (\url{https://github.com/mortamini/vbnet}).}

\textcolor{black}{\section*{Acknowledgements}}
\textcolor{black}{The authors would like to thank the two anonymous reviewers for their constructive comments and suggestions, which have significantly improved the quality of this manuscript. Mohammad Arashi’s work is based on the research supported in part by the Iran National Science Foundation (INSF) grant No. 4015320.}

\bibliographystyle{apa}
{

}

\end{document}